\documentclass[10pt,twocolumn,letterpaper]{article}

\usepackage{iccv}              

\definecolor{iccvblue}{rgb}{0.21,0.49,0.74}
\usepackage[pagebackref,breaklinks,colorlinks]{hyperref}
\hypersetup{
    linkcolor=iccvblue,    
    citecolor=iccvblue,    
    filecolor=iccvblue,    
    urlcolor=magenta       
}

\usepackage{graphicx}

\usepackage{marvosym}

\title{CineMaster: A 3D-Aware and Controllable Framework for Cinematic Text-to-Video Generation\vspace{-10pt}}

\author{\textbf{Qinghe Wang}$^{1*}$ \  \textbf{Yawen Luo}$^{2*}$ \ \textbf{Xiaoyu Shi}$^{3\dagger}$ \  \textbf{Xu Jia}$^{1}\textsuperscript{\Letter}$ \ \textbf{Huchuan Lu}$^{1}$ \  \textbf{Tianfan Xue}$^{2}\textsuperscript{\Letter}$\\ 
\textbf{Xintao Wang}$^{3}$ \ \textbf{Pengfei Wan}$^{3}$ \   \textbf{Di Zhang}$^{3}$ \ \textbf{Kun Gai}$^{3}$\\
$^1$Dalian University of Technology \quad $^2$The Chinese University of Hong Kong \quad $^3$Kuaishou Technology\\
$^*$Equal contribution \quad $^\dagger$Project Leader \quad $\textsuperscript{\Letter}$Corresponding author\vspace{0.3cm}\\
\textcolor{magenta}{\url{https://cinemaster-dev.github.io/}}
}

\begin{document}

\twocolumn[{%
\renewcommand\twocolumn[1][]{#1}%
\maketitle
\vspace{-0.9cm}
\begin{center}
    \centering
    \captionsetup{type=figure}
    \includegraphics[width=1\linewidth]{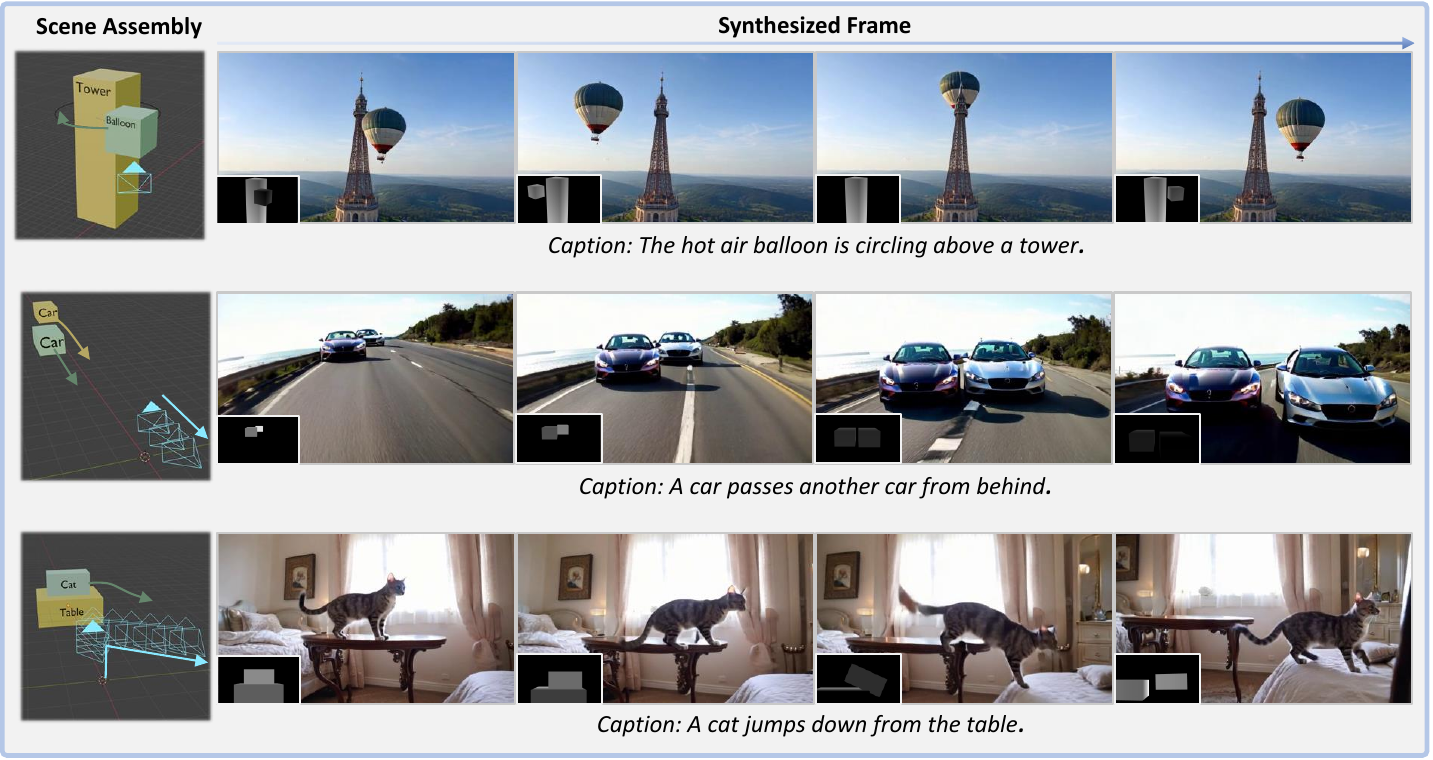}
    \vspace{-12pt}
    \captionof{figure}{CineMaster targets at granting users 3D-aware and intuitive control over the text-to-video generation process. We first design a 3D-native workflow that enables users to manipulate objects and camera in the 3D space. Then the rendered depth maps and camera trajectories serve as strong guidance to synthesize the desired video content. Left column shows the objects and camera setup using the proposed workflow. Right columns indicate synthesized frames with rendered depth maps on the bottom left.}
    \label{fig:teaser}
    \vspace{3pt}
\end{center}%
}]

\newcommand\blfootnote[1]{%
\begingroup
\renewcommand\thefootnote{}\footnote{#1}%
\addtocounter{footnote}{-1}%
\endgroup
}

\blfootnote{$^*$Work done during internship at KwaiVGI, Kuaishou Technology.}


\begin{abstract}
In this work, we present CineMaster, a novel framework for 3D-aware and controllable text-to-video generation. Our goal is to empower users with comparable controllability as professional film directors: precise placement of objects within the scene, flexible manipulation of both objects and camera in 3D space, and intuitive layout control over the rendered frames. To achieve this, CineMaster operates in two stages.
In the first stage, we design an interactive workflow that allows users to intuitively construct 3D-aware conditional signals by positioning object bounding boxes and defining camera movements within the 3D space. In the second stage, these control signals—comprising rendered depth maps, camera trajectories and object class labels—serve as the guidance for a text-to-video diffusion model, ensuring to generate the user-intended video content. Furthermore, to overcome the scarcity of in-the-wild datasets with 3D object motion and camera pose annotations, we carefully establish an automated data annotation pipeline that extracts 3D bounding boxes and camera trajectories from large-scale video data. Extensive qualitative and quantitative experiments demonstrate that CineMaster significantly outperforms existing methods and implements prominent 3D-aware text-to-video generation. 
\end{abstract}

\section{Introduction}
\label{sec:intro}
With the advances of diffusion models~\cite{ddpm,ddim,sd} and large-scale pretraining paradigms, text-to-video generation (T2V)~\cite{VideoComposer,chen2023videocrafter1,svd,animatediff} has experienced rapid development. The T2V models empower both artists and novices to create impressive videos merely by providing textual prompts. Subsequently, controllable video generation has gained increasing attention, driven by the growing demand for fine-grained control over the video creation process.

Ideally, a controllable T2V framework should grant users comparable controllability as a professional film director: allowing precise placement of objects within a scene, flexible manipulation of both objects and camera, and intuitive layout control over each rendered frame.
However, existing approaches fall short of achieving this vision.
Early works typically extend image-based ControlNet~\cite{controlnet,mou2024t2i} to the video domain, guiding generation using condition maps (e.g. depth~\cite{chen2023control}, semantic~\cite{wang2024easycontrol}, optical flow~\cite{motioni2v,koroglu2024onlyflow,bian2025gs}, or canny edge maps~\cite{wang2024easycontrol,guo2023sparsectrl}). Yet, these methods generally rely on pre-existing videos to obtain condition maps, since it is non-trivial to create such condition maps from scratch. Moreover, in terms of controllability, MotionCtrl~\cite{MotionCtrl} and Direct-A-Video~\cite{yang2024direct} offer preliminary control over both object and camera movements, but these methods only allow 2D object control which is very different from how filmmakers or video creators plan a shooting in 3D space. More recent works have moved towards integrating 3D-aware signals into video generation~\cite{fu20243dtrajmaster,chen2025perception,shader,shuai2025free}. However, these methods are only designed for image-to-video generation or rely on synthetic data from Unreal Engine.

\begin{figure}[!t]
  \centering
  \includegraphics[width=\linewidth]{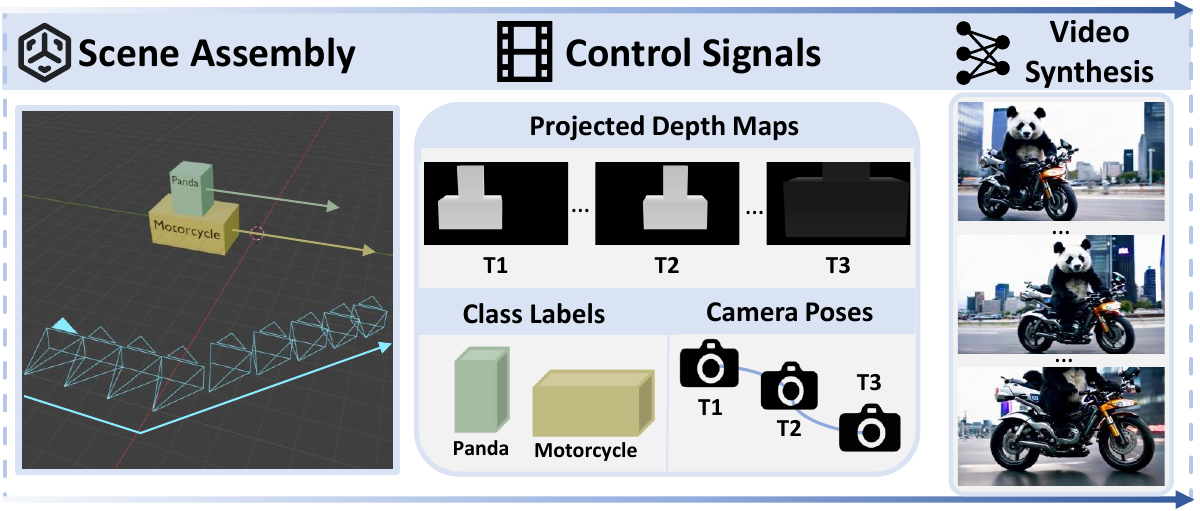}
  \vspace{-15pt}
  \caption{Overview of CineMaster. CineMaster consists of two stages. First, we present an interactive workflow that allows users to intuitively manipulate the objects and camera in a 3D-native manner. Then the control signals are rendered from the 3D engine and fed into a text-to-video diffusion model, guiding the generation of user-intended video content.}
  \vspace{-8pt}
  \label{fig:overview}
\end{figure}

To bridge the aforementioned gap, we present CineMaster, a framework designed for highly controllable text-to-video generation as shown in Fig.~\ref{fig:teaser}. Specifically, CineMaster operates in two stages, as presented in Fig.~\ref{fig:overview}. The first stage is an interactive workflow that allows users to specify the requirements (conditions) of a generated video, similar to how filmmakers design a capturing plan. It allows users to describe the primary objects in a scene using a set of 3D bounding boxes with semantic labels. These bounding boxes, along with the camera, can be repositioned across keyframes, allowing users to orchestrate complex motion dynamics. After each modification, CineMaster provides a preview of the rendered frames for iterative refinement until the desired rendered effects are achieved. In the second stage, we finetune a text-to-video diffusion model to generate a video conditioned on the control signals provided in the first stage. Crucially, in addition to camera trajectories and user-provided class labels, we propose to utilize the rendered depth maps of all frames as augmented visual cues. These depth maps explicitly contain the desired 3D layout of each frame, serving as strong guidance for the diffusion model to generate the user-intended video content.

One challenge for this design is the lack of videos with ground-truth 3D bounding box and camera trajectory annotations. To solve this limitation, we further propose an automatic data labeling pipeline, as illustrated in Sec~\ref{sec:dataset_labeling_pipeline}. Using this pipeline, we build the largest video datasets with both the ground-truth 3D bounding box and 3D camera trajectory annotations.

Finally, to evaluate the controllability of our proposed framework, we conduct extensive experiments, comparing it with existing SOTA methods, and performing ablative studies to validate the effectiveness of our core modules.

\section{Related Work}
\label{sec:formatting}

\noindent\textbf{Controllable Video Generation via Planar Condition Maps.} The pioneer works ControlNet~\cite{controlnet} and T2I-Adapter~\cite{mou2024t2i} introduce the paradigm of conditioning generation on planar maps in the image generation field. Subsequently, many works extend this paradigm to the video domain using different condition maps, e.g. depth maps~\cite{chen2023control}, human pose maps~\cite{hu2024animate}, semantic maps~\cite{wang2024easycontrol} and optical flow maps~\cite{motioni2v,koroglu2024onlyflow,bian2025gs}. However, they generally assume the existence of such condition maps, but it is indeed non-trivial to create precise condition maps from scratch, especially for novices. Therefore, we carefully design an interactive workflow to help users obtain 3D-aware condition maps in an intuitive way. We also get inspiration from LooseControl~\cite{bhat2024loosecontrol}, to use 3D bounding box as an appropriate abstract representation of objects in the scene.

\begin{figure*}[!t]
  \centering
  \includegraphics[width=\linewidth]{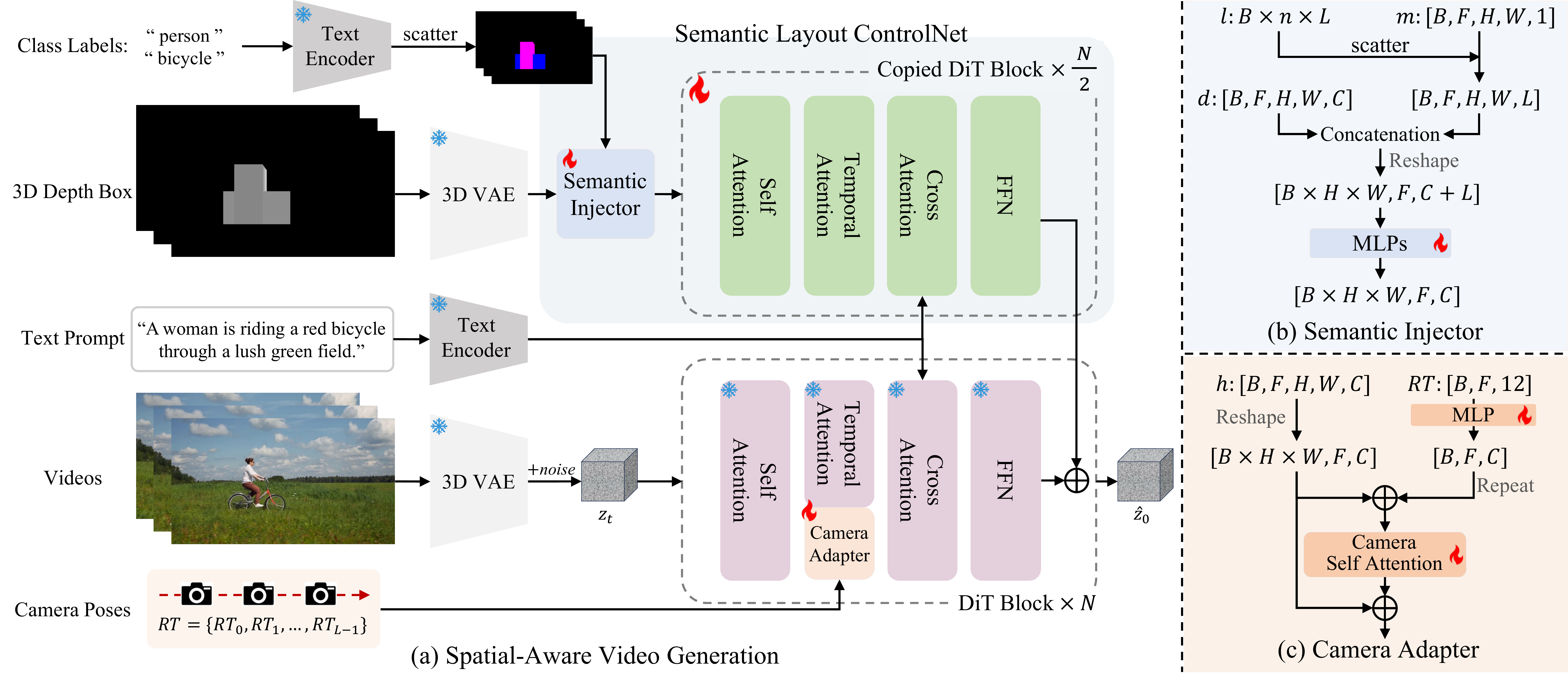}
  \vspace{-15pt}
  \caption{Overview of the network architecture. We design a Semantic Layout ControlNet which consists of a semantic injector and a DiT-based ControlNet. Semantic injector fuses the 3D spatial layout and class label conditions. The DiT-based ControlNet further represents the fused features and adds to the hidden states of the base model. Meanwhile, we inject the camera trajectories by the camera adapter to achieve joint control over object motion and camera motion.}
  \label{fig:network}
  \vspace{-4pt}  
\end{figure*}

\noindent\textbf{Object Motion Control.} Previous methods primarily focus on motion control in 2D space. MotionCtrl~\cite{MotionCtrl}, DragNUWA~\cite{dragnuwa}, and Tora~\cite{Tora2024} represent object motion trajectories as sequences of spatial positions, encoding coordinates into dense control maps. Additionally, 2D bounding boxes have been adopted as control signals to enable flexible motion generation in Direct-A-Video~\cite{yang2024direct} and Boximator~\cite{boximator}. Motion control through sketches has also been explored in VideoComposer~\cite{VideoComposer}. 
While these methods have demonstrated capabilities in object motion control, their control signals limit the controllability only in 2D space. 3DTrajMaster~\cite{fu20243dtrajmaster} is the first to use 6D pose sequences of objects to control object motion in 3D space.

\noindent\textbf{Camera Motion Control.} Camera pose serves as a crucial control signal in video generation, determining which portions of the scene are captured and presented in the final output. MotionCtrl~\cite{MotionCtrl} pioneers the integration of camera poses as control signals for camera movement manipulation. Building upon this foundation, CameraCtrl~\cite{cameractrl} introduces the use of Plücker embeddings of camera poses to enhance motion controllability. These methods are all trained on an indoor dataset RealEstate10K~\cite{realestate10k} for learning camera motion which limits the ability to generalize to in-the-wild scenes. Direct-A-Video~\cite{yang2024direct} uses data augmentations on static videos to simulate basic camera movements~(only pan and zoom movements) and employs Fourier embedder and temporal cross-attention layers to inject camera poses. It cannot generalize to more complex camera movements such as ``Anti-clockwise". Therefore, the development of this field is constrained by the scarcity of large-scale in-the-wild datasets with camera pose annotations.

\noindent\textbf{Joint Motion Control.} Based on the preliminary explorations of joint motion control~\cite{MotionCtrl,yang2024direct}, some concurrent works further advance this field. Motion Prompting~\cite{geng2024motion} leverages 2D point tracking results to represent object motion and camera motion. Perception-as-Control~\cite{chen2025perception} and DaS~\cite{shader} further capture 3D point tracking results by SpatialTracker~\cite{SpatialTracker} to extend the joint control into 3D space. The former denotes objects of reference image as unit spheres with different colors. The latter directly uses the point tracking video as the motion condition. MotionCanvas~\cite{motioncanvas}
also measures camera motion by point tracking and renders 2D instance box map as object global motion representation. However, these methods are all designed for the image-to-video generation task which only animates an initial image and cannot plan a shooting in 3D space from scratch. SynFMC~\cite{shuai2025free} uses Unreal Engine to render both 6D pose of objects and camera to construct video datasets for joint motion control, but the limited diversity and the domain gap of UE data restrict the model's generalizability. To overcome the scarcity of in-the-wild datasets with both 3D object motion and camera pose annotations, we carefully establish an automated data annotation pipeline to extract 3D bounding boxes and camera trajectories from large-scale video data for learning joint motion control. In addition, we provide an interactive workflow to allow users to intuitively manipulate objects and camera in 3D scene.

\section{Method}
Controllable text-to-video generation (T2V) targets at providing more conditional guidance beyond textual prompts, thereby enabling fine-grained control over the video generation process. We present CineMaster, which aspires to give users a level of controllability comparable to professional film directors: allowing precise object placement within a scene, flexible manipulation of both objects and camera, and intuitive layout control over each rendered frame. Our proposed CineMaster operates in two stages. In the first stage~(Sec~\ref{sec:method_stage_1}), we propose an interactive workflow for constructing 3D-aware control signals. In the second stage (Sec~\ref{sec:method_stage_2}), these control signals serve as conditions for T2V models to synthesize the desired video content. Moreover, due to the scarcity of large-scale datasets with 3D bounding box and camera trajectory annotations, we carefully develop an automated data labeling pipeline (Sec~\ref{sec:dataset_labeling_pipeline}).

\subsection{Stage 1: 3D-Aware Control Signals}
\label{sec:method_stage_1}
The first stage centers on constructing 3D-aware control signals through a user-friendly workflow. Inspired by LooseControl~\cite{bhat2024loosecontrol}, we employ 3D bounding boxes as the principal form of object representation. Users can freely adjust the size and position of these bounding boxes within the 3D scene. By repositioning bounding boxes and camera across keyframes, users gain intuitive control over object and camera trajectories, effectively dictating the motion dynamics. Another key component of our workflow is the preview mechanism, which lets users examine rendered frames after each modification.

This workflow closely mirrors real-world filmmaking: directors typically arrange actor and camera movements in multiple takes, reviewing footage on monitors to refine the final shots. Once satisfactory rendering effects are achieved, we export camera trajectories and per-frame projected depth maps for use in the subsequent stage. The primary advantage of this system is its 3D-native and intuitive nature. We implement the interactive system using the open-source engine Blender, where users select keyframes for object and camera placement. The system then automatically interpolates trajectories for intermediate frames, providing a seamless and efficient workflow for complex scene setup.

\subsection{Stage 2: Conditional Video Generation}
\label{sec:method_stage_2}
We condition a base text-to-video model on the control signals derived from the first stage. Crucially, beyond using the camera trajectory and object labels as inputs, we also introduce projected depth maps of each frame as augmented visual condition. These depth maps explicitly encode the desired 3D layout, providing strong guidance for the diffusion model to generate accurate video content. To effectively integrate these additional inputs into the T2V model, we design two key components: a semantic layout injection module and camera adapter, as illustrated in Fig.~\ref{fig:network}.

\noindent\textbf{Base Model.} Our model is developed upon a pretrained text-to-video foundation model, which consists of a 3D Variational Auto-Encoder (VAE)~\cite{kingma2013auto}, T5 encoder~\cite{raffel2020exploring} and a transformer-based latent diffusion model~\cite{peebles2023scalable,chen2023pixart}. Each basic transformer block is instantiated as a sequence of 2D spatial self-attention, 3D spatial-temporal self-attention, text cross attention and feed-forward network~(FFN). The text prompts are encoded as $c_{text}$ by T5 encoder to guide the generation model. We define a straight forward path between clean data $z_0$ and noised data $z_t$ at timestep $t$ with Rectified Flow~\cite{esser2024scaling}:
\begin{equation}
\label{add_noise}
     z_t = (1-t)z_0 + t\epsilon,
\end{equation}
where $\epsilon\in\mathcal{N}(0,\mathbf{I})$. The denoising process is defined as a mapping from $z_t$ to $z_0$ by an ordinary differential equation (ODE):
\begin{equation}
\label{denoise}
     dz_t = v_{\Theta}(z_t,t,c_{text})dt,
\end{equation}
where the velocity $v$ is parameterized by the weights $\Theta$ of the denoising network. The training process is supervised by Conditional Flow Matching~\cite{lipman2022flow} to regress velocity:
\begin{equation}
\label{lcm_loss}
     \mathcal{L}_{LCM} = \mathbb{E}_{t,\epsilon\sim\mathcal{N}(0,\mathbf{I}),z_0}\left[\|(z_1-z_0)-v_{\Theta}(z_t,t,c_{text})\|^2_2\right]
\end{equation}

\noindent\textbf{Semantic Layout ControlNet.} We design a Semantic Layout ControlNet to integrate the 3D spatial layouts from the projected depth maps with semantic information from per-entity class labels. Specifically, we copy $N/2$ DiT blocks from the base model to form the DiT-based ControlNet. Projected depth maps could represent the spatial layout of a scene, but when multiple subjects appear in text prompts, it is difficult to specify the position of each entity only by text prompts. Therefore, we use text encoder to represent $n$ entity class labels of the input video as text embeddings~$l\in\mathbb{R}^{n\times L}$, and fuse these embeddings to the corresponding position specified by the downsampled entity mask~$m\in\mathbb{R}^{F\times H\times W\times 1}$ as shown in Fig.~\ref{fig:network}(b). $L$ denotes the dimension of a text embedding. $H$ and $W$ denote the spatial size of the latent encoded by VAE, and $C$ is the number of channels. The 3D VAE latents of projected depth maps~$d\in\mathbb{R}^{F\times H\times W\times C}$ are concatenated with the semantic embeddings along channel dimension and fused by MLPs. The hidden states of ControlNet are incorporated into the base model to provide semantic layout guidance for the generation process.

\begin{figure*}[!t]
  \centering
\includegraphics[width=\linewidth]{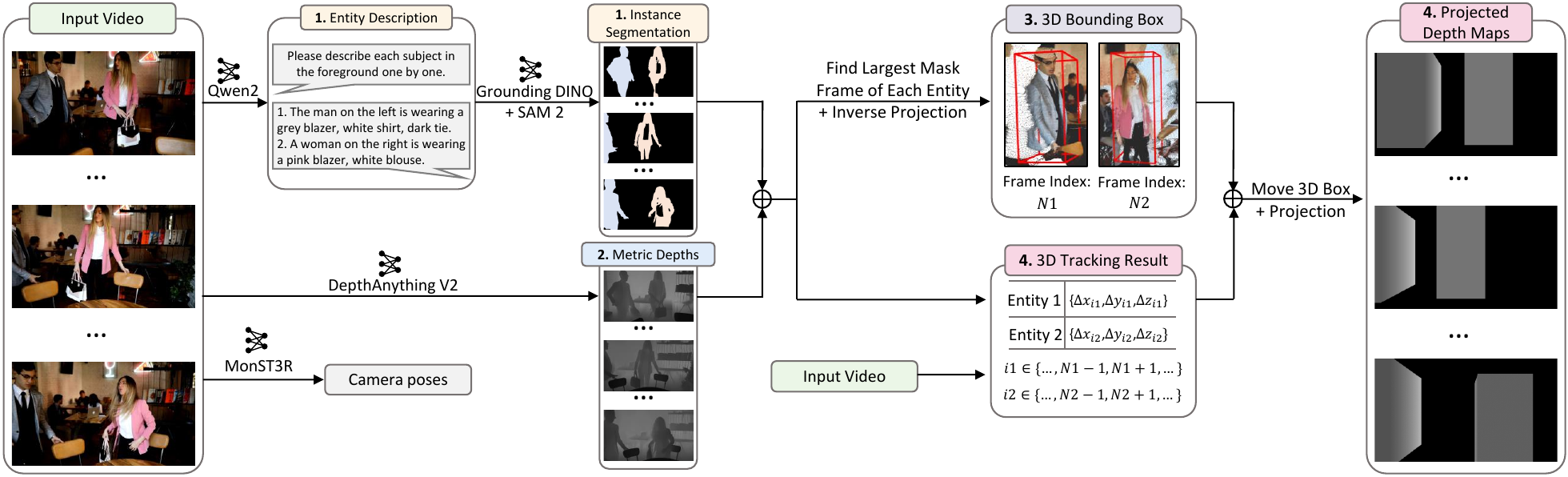}
  \caption{Dataset Labeling Pipeline. We propose a data labeling pipeline to extract 3D bounding boxes, class labels and camera poses from videos. Our pipeline consists of four steps: 1) Instance Segmentation: Obtain instance segmentation results from the foreground in videos. 2) Depth Estimation: Produce metric depth maps using DepthAnything V2. 3) 3D Point Cloud and Box Calculation: Identify the frame with the largest mask for each entity and compute the 3D point cloud of each entity through inverse projection. Then, use the minimum volume method to calculate the 3D bounding box for each entity. 4) Entity Tracking and 3D Box Adjustment: Access the point tracking results of each entity and calculate the 3D bounding boxes for each frame. Finally, project the entire 3D scene into depth maps.}
  \label{fig:dataset_pipeline}
  \vspace{-8pt}

\end{figure*}

While the proposed Semantic Layout ControlNet enables precise control over the 3D position of each entity in generated videos, relying exclusively on 3D bounding boxes might introduce ambiguity between object and camera movements. For instance, if the bounding box of a balloon shifts upward, it could indicate that the ballon is rising, the camera is moving downward, or both. To resolve this ambiguity, we additionally inject explicit camera poses into the generation process, allowing the model to distinguish object motion from camera trajectories more reliably.

\noindent\textbf{Camera Adapter.} In each DiT block, we incorporate the camera condition via a residual connection situated between the self-attention module and temporal-attention module as shown in Fig.~\ref{fig:network}(c). Specifically, each camera pose is represented by a $3\times 3$ rotation matrix and a $3\times 1$ translation matrix. We take a sequence of camera pose $RT=\{RT_0,RT_1,\cdots,RT_{F-1}\}\in\mathbb{R}^{F\times 12}$ as input, where $F$ is the length of latent frames. An MLP first aligns the dimension of $RT$ to the token length $C$. We then add the camera pose features to each token in the hidden states. These fused hidden states are fed into the self-attention module to inject camera motion information. In addition, the camera pose features are introduced again via the residual connection. By leveraging this camera adapter, the proposed CineMaster could support joint control over object motion and camera motion.

\subsection{Dataset Labeling Pipeline}
\label{sec:dataset_labeling_pipeline}

There is a lack of large-scale video datasets with 3D bounding box and camera pose annotations. To train the second-stage network, we design an automated data labeling pipeline as shown in Fig.~\ref{fig:dataset_pipeline}. It takes an in-the-wild video as input and extracts the required class labels, camera trajectories and projected depth maps.

\noindent\textbf{Class Labels.} To obtain the class labels for objects present in the scene, we perform instance segmentation for each entity in the video. To achieve open-set instance segmentation, we combine Grounding DINO\cite{groundingdino} with SAM 2~\cite{2024sam2segmentimages}, where Grounding DINO produces $2$D bounding boxes guided by entity descriptions. To enhance foreground entity detection, we utilize the multi-modal large model Qwen2~\cite{qwen2} to generate entity descriptions as guidance for Grounding DINO. This process yields $2$D bounding boxes and class labels for each entity in the first frame, which then guides SAM 2 for video segmentation. To address potential issues with overlapping boxes and incorrect class labels from Grounding DINO, we implement crucial post-processing steps: a box IOU filter and feature similarity verification between the regions within $2$D boxes and their assigned class labels.

\noindent\textbf{Camera Trajectories.} We employ the SOTA camera pose estimation model MonST3R~\cite{zhang2024monst3r} to obtain camera trajectories throughout the video sequence.

\noindent\textbf{Projected Depth Maps.} We employ DepthAnything V2~\cite{depth_anything_v2} to generate metric depth maps for the entire video sequence, which are essential for the subsequent inverse projection process. 
The third step involves inverse projection to obtain $3$D boxes for each entity. We operate under the assumption that each entity maintains a constant volume in the $3$D scene. To address cases where entities may appear partially in certain frames, we identify the optimal frame index for each entity, typically when the entity is most completely visible, to ensure accurate inverse projection and adequate volume representation. For each entity, we combine the instance segmentation mask with the corresponding metric depth map at this optimal frame to generate a $3$D point cloud, from which we derive the minimal-volume $3$D bounding box.

\begin{figure*}[!t]
  \centering
  \includegraphics[width=\linewidth]{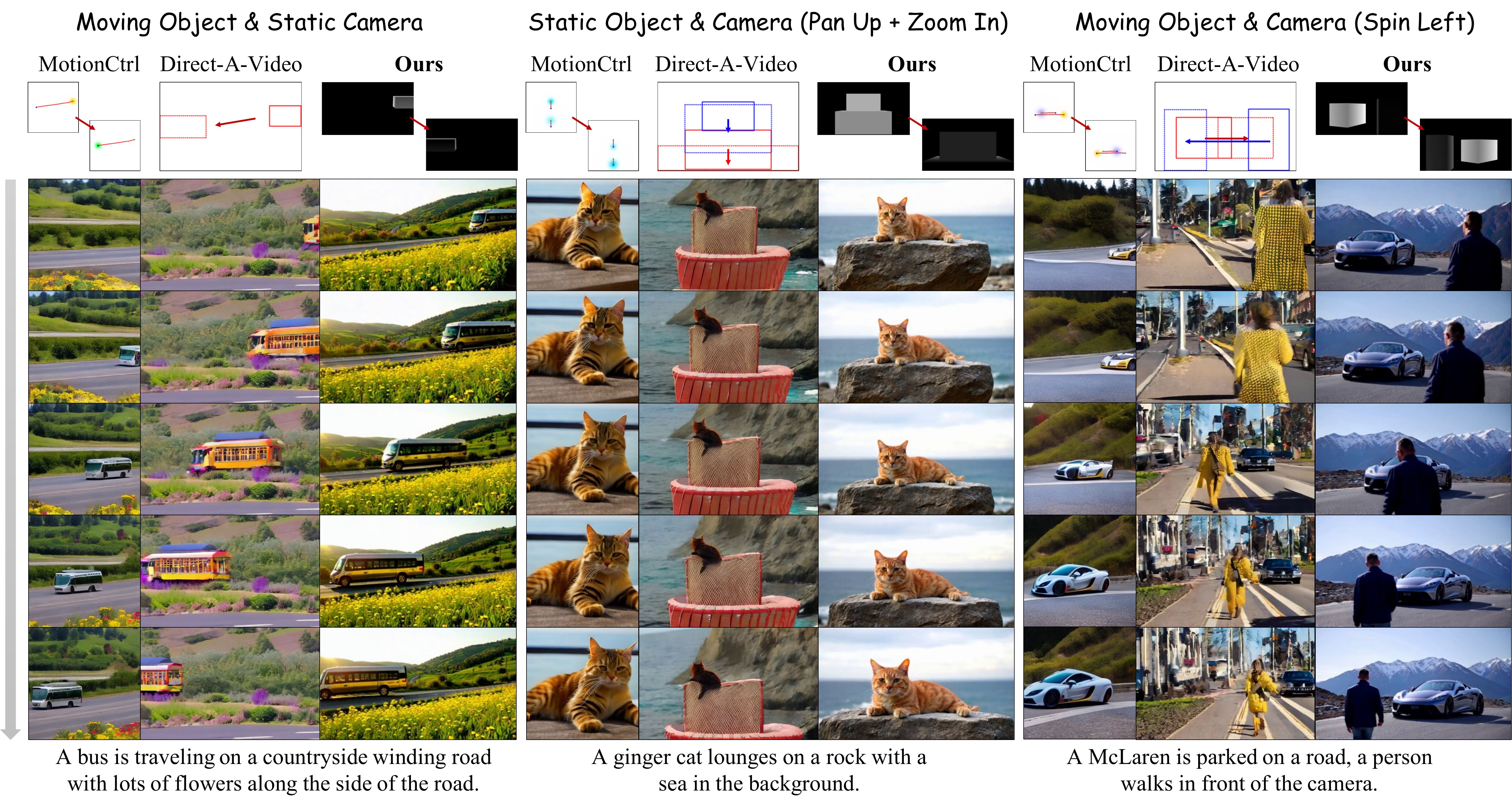}
  \vspace{-20pt}  
  \caption{We present three different feature comparisons: moving object $\&$ static camera, static object $\&$ moving camera and moving object $\&$ moving camera. We transform our 3D box condition to object trajectories for MotionCtrl~\cite{MotionCtrl} and 2D bounding box sequences for Direct-A-Video~\cite{yang2024direct} to align the input conditions. In comparison, CineMaster could better control object motion and camera motion separately or jointly to generate diverse user-intended scenes.}
  \label{fig:comparison}
  \vspace{-8pt}  
\end{figure*}

Following the establishment of maximum 
$3$D boxes for all entities, the final step involves computing temporal-spatial transformations of these boxes within the $3$D scene for all frames. 
We conduct 3D point tracking by SpatialTracker~\cite{SpatialTracker} starting from the optimal frame $N(j)$ of $j$th object to the rest frames $\{..., N(j)-1, N(j)+1, ...\}$, and the average inter-frame displacements $\{\Delta x_{ij}, \Delta y_{ij}, \Delta z_{ij}\}$ of all feature points from each object are regarded as the spatial movement of the $j$th object's 3D box where $ij\in\{..., N(j)-1, N(j)+1, ...\}$ denotes $i$th frame of $j$th object. Then we can compute the 3D boxes of all objects in all frames. To represent 3D boxes as explicit control signals, we further project the constructed 3D boxes into image space and render depth maps.

\section{Experiments}
\subsection{Experimental Setup}
\noindent\textbf{Training Paradigm.} We design a dedicated training strategy consisting of three stages, i.e., 1) training a DiT-based ControlNet on dense depth maps, 2) adapting ControlNet to 3D box datasets, and 3) jointly training Semantic Layout ControlNet and Camera Adapter. Specifically, following LooseControl~\cite{bhat2024loosecontrol}, we first train our DiT-based ControlNet on 167K videos crawled from the Internet with dense depth maps labeled by DepthAnything V2~\cite{depth_anything_v2}. Subsequently, we use our data annotation pipeline to construct a 3D box dataset with 156K videos and 118K images for training Semantic Layout ControlNet. The images are collected from COCO~\cite{lin2014microsoft} and Object365~\cite{shao2019objects365} which provide more categories and precise instance segmentation annotations. With the metric depth maps measured by DepthAnything V2~\cite{depth_anything_v2}, we could obtain the 3D boxes of image datasets by calculating the 3D box with minimal volume for each object in the point cloud. By the image-video joint training, we integrate the spatial layout and semantic information into the Semantic Layout ControlNet which could guide the base model to generate box-aligned videos with specified class labels. We further annotate 99.6K videos out of the 156k videos using our proposed pipeline to obtain camera poses. We also utilize RealEstate10K~\cite{realestate10k} dataset which features larger camera motion, resulting in 10.4K data samples. We sample data between our dataset and Real-Estate10K with 3:1 ratio for training to enhance the learning of larger camera motion. 
Based on the merged video dataset with both 3D box and camera pose annotations, we train the Semantic Layout ControlNet and Camera Adapter jointly and master the joint controllability of object motion and camera motion for flexible customized video generation.

\noindent\textbf{Implementation Details.} We train CineMaster based on our internal text-to-video generation model with $\sim1B$ parameters for research purposes. Following NaViT~\cite{dehghani2024patch}, we pad the videos to the same shape for each batch managed by attention masks during training. Each training video segment contains 77 frames~(i.e., 5 seconds) sampled with 15 frames per second~(fps). We use Adam optimizer~\cite{kingma2014adam} and train on 24 NVIDIA A800 GPUs, with a batch size of 4 and a learning rate of $5\times10^{-5}$. The three stages of the training process consist of 12,000, 7,000, 6000 steps respectively. During inference, we set the scale of classifier-free guidance~\cite{ho2022classifier} as 12.5 and the DDIM~\cite{song2020denoising} steps as 50. We make a trade-off between object motion and camera motion by injecting semantic layout information and camera poses with 25 and 15 steps respectively.

\noindent\textbf{Baselines.} We compare CineMaster with existing SOTA methods MotionCtrl~\cite{MotionCtrl} and Direct-a-Video~\cite{yang2024direct} which could also control object motion and camera motion simultaneously. To align with different input requirements, we convert our 3D box condition into object trajectories for MotionCtrl and 2D bounding box sequences for Direct-A-Video. In addition, we empirically align the coordinate system and scale of the input camera poses for comparison.

\begin{table}[!t]
\centering
\caption{Quantitative comparisons with baselines. $\uparrow$ indicates higher is better, while $\downarrow$ indicates that lower is better. The best result is shown in \textbf{bold}. Our CineMaster outperforms previous SOTA baselines on all metrics.
 }
\vspace{-4pt}
\setlength{\tabcolsep}{1.5pt}
\label{tab:comparsion}
\begin{tabular}{lccccc}
\toprule
               & mIoU$\uparrow$           & Traj-D$\downarrow$        & FVD$\downarrow$               & FID$\downarrow$             & CLIP-T$\uparrow$              \\
\midrule                
MotionCtrl     & $-$          & 94.82          & 2163.0          & 201.6          & 0.302                    \\
Direct-A-Video & 0.332          & 83.53          & 1966.3          & 183.5          & 0.273                   \\
Ours           & \textbf{0.551} & \textbf{66.29} & \textbf{1530.9} & \textbf{175.9} & \textbf{0.321}  \\
\bottomrule
\vspace{-18pt}
\end{tabular}
\end{table}

\noindent\textbf{Evaluation Metrics.}
1) \textit{Object-box alignment}: We use Grounding DINO~\cite{groundingdino} to detect 2D object boxes in generated videos to measure the mean Intersection over Union~(mIoU) and measure the trajectory deviation~(Traj-D) by calculating the difference of the center points against ground truths. In addition, to evaluate the depth control accuracy of the generated objects, we calculate the average depth of the object regions in each generated frame using SAM 2~\cite{2024sam2segmentimages} and DepthAnything V2~\cite{depth_anything_v2}, and measure the depth deviation~(Depth-D) by the Root Mean Squared Error (RMSE) with the depth values of the given 3D boxes. 2) \textit{Video quality}: We employ Fréchet Video Distance (FID)~\cite{unterthiner2018towards}, Fréchet Inception Distance (FID)~\cite{lucic2017gans} and CLIP Similarity~(CLIP-T) to evaluate the appearance of generated results.

\subsection{Comparison with Other Methods}
\noindent\textbf{Qualitative Comparison.} As shown in Fig.~\ref{fig:comparison}, we show three different feature comparisons: moving object $\&$ static camera, static object $\&$ moving camera and moving object $\&$ moving camera. In the first setting, MotionCtrl~\cite{MotionCtrl} moves the camera rightward to align the object trajectory, but fails to either maintain the camera stationary or make the object move. It shows that there is still the camera motion and object motion coupling issue in MotionCtrl. In the third setting, since MotionCtrl is unable to associate multiple trajectories with their respective objects, it generates the ``McLaren" that appears to follow the trajectory of the ``person" and fails to generate the ``person". Direct-A-Video~\cite{yang2024direct} presents low-quality textures for generating ``bus" and ``rock" which demonstrates that its control disturbs the generation quality. It exhibits weaker camera movement and box alignment, and produces unexpected shot changes and more artifacts in the third setting. In comparison, the proposed CineMaster performs the best control performance for the control of object motion and camera motion in three settings.

\noindent\textbf{Quantitative Comparison.} In addition, we further report the quantitative comparison in Table~\ref{tab:comparsion}. Since MotionCtrl only uses point trajectory sequences to specify the positions of generated objects and ignores the spatial size, we do not calculate its mIoU. It does not explicitly associate multiple trajectories with their respective objects and suffers from the camera motion and object motion coupling issue. Therefore, it obtains unsatisfactory Traj-D. Direct-A-Video only trains for learning camera movement, and controls the object motion by spatial cross-attention modulation with associated object words and box trajectories to guide the spatial-temporal placement of objects only for inference. The training-free object motion control biases the vanilla inference distribution, resulting in the degradation of generation quality. In addition, no joint training of object motion and camera motion control leads to a gap between training and inference, so it obtains weak mIoU and Traj-D. In contrast, we construct the video dataset with both 3D box and camera pose for joint training, and the proposed CineMaster could harmoniously control the object motion and camera motion simultaneously. Therefore, CineMaster outperforms previous SOTA methods on all metrics. In particular, we achieve significantly higher mIoU and Traj-D, indicating that our framework can generate videos that better follow the user's spatial design.

\begin{table}[t]
\centering
\caption{Ablation study for training paradigms. Details of each setting are introduced in Sec~\ref{sec:ablation}. Overall, the setting of ``Joint Train"~(our final version) achieves the best performance on all metrics than other variants.}
\vspace{-4pt}
\setlength{\tabcolsep}{1.5pt}
\label{tab:ablation}
\resizebox{\linewidth}{!}{
\begin{tabular}{lcccccc}
\toprule
 & mIoU$\uparrow$           & Traj-D$\downarrow$        & FVD$\downarrow$               & FID$\downarrow$             & CLIP-T$\uparrow$      & Depth-D$\downarrow$      \\
\midrule                        
w/o stage 1   & 0.544          & 72.27          & 176.4          & 1576.1          & 0.310          & 0.725          \\
w/o semantic  & 0.391          & 83.81          & 177.4          & 1622.2          & 0.304          & 0.717          \\
Isolated S,C  & 0.480          & 70.53          & 180.2          & 1840.9          & 0.313          & 0.705          \\
Fix S train C & 0.545          & 68.15          & 177.8          & 1673.9          & 0.317          & 0.702          \\
Joint Train   & \textbf{0.551} & \textbf{66.29} & \textbf{175.9} & \textbf{1530.9} & \textbf{0.321} & \textbf{0.685}\\
\bottomrule
\end{tabular}}
\vspace{-13pt}
\end{table}

\subsection{Ablation Study}
\label{sec:ablation}
We experiment with different training paradigms to validate the effectiveness of the delicate designs in our workflow:
\begin{itemize}
\item ``w/o stage 1": without training the DiT-based ControlNet on dense depth maps, the training process starts directly from the second training stage.
\item ``w/o semantic": without semantic injector, this setting does not specify the class labels of 3D boxes.
\item ``Isolated S,C": Semantic Layout ControlNet and Camera Adapter are trained separately and used together for inference.
\item ``Fix S train C": this setting first trains Semantic Layout ControlNet to convergence, then freezes its weights and trains the Camera Adapter. 
\item ``Joint Train"~(our final version): Semantic Layout ControlNet and Camera Adapter are trained simultaneously.
\end{itemize}

As shown in Table~\ref{tab:ablation}, the setting of ``w/o stage 1" lacks the fine-grained perception for depth control signal, so it obtains mediocre Depth-D. The positions of the generation objects can only be specified via text prompt in the setting of ``w/o semantic", resulting in the poor mIoU, Traj-D and CLIP-T. The setting of ``Isolated S,C" faces a discrepancy between training and inference phases, as the Semantic Layout ControlNet and Camera Adapter are trained separately without cross-module communication, resulting in degraded generation quality supported by lower FVD and FID. Although the setting of ``Fix S train C" trains the camera adapter based on the frozen Semantic Layout ControlNet, it still fails to eliminate the coupling of camera motion and object motion already learned in the frozen Semantic Layout ControlNet, leading to suboptimal FVD and FID. 
Benefiting from the constructed video dataset with 3D box and camera pose labels, we experimentally observe that the setting of ``Joint Train" could harmoniously integrate the control for camera motion and object motion and perform the best results on all metrics.

\section{Limitations and Conclusions}
Ideally, 3D bounding box can naturally and precisely control the orientation of objects in space. For instance, when we rotate the 3D box of a human, it should produce a video sequence of a human turning around. However, the community currently lacks accurate open-set object pose estimation models. Therefore, we leave this promising functionality as future work. In conclusion, our project stems from the goal of granting users the creation controllability as professional film directors. To this end, we propose CineMaster for highly controllable text-to-video generation. Specifically, we first design a 3D-native workflow that allows users to manipulate objects and camera in an intuitive manner. Then we train a conditional text-to-video diffusion model to synthesize the user-intended videos. We emphasize the importance of adopting projected depth maps as strong visual control signals. Extensive experiments demonstrate that CineMaster achieves controllable and 3D-aware cinematic video generation.


\begin{thebibliography}{50}
\providecommand{\natexlab}[1]{#1}
\providecommand{\url}[1]{\texttt{#1}}
\expandafter\ifx\csname urlstyle\endcsname\relax
  \providecommand{\doi}[1]{doi: #1}\else
  \providecommand{\doi}{doi: \begingroup \urlstyle{rm}\Url}\fi

\bibitem[Bhat et~al.(2024)Bhat, Mitra, and Wonka]{bhat2024loosecontrol}
Shariq~Farooq Bhat, Niloy Mitra, and Peter Wonka.
\newblock Loosecontrol: Lifting controlnet for generalized depth conditioning.
\newblock In \emph{ACM SIGGRAPH 2024 Conference Papers}, pages 1--11, 2024.

\bibitem[Bian et~al.(2025)Bian, Huang, Shi, Li, Wang, and Li]{bian2025gs}
Weikang Bian, Zhaoyang Huang, Xiaoyu Shi, Yijin Li, Fu-Yun Wang, and Hongsheng Li.
\newblock Gs-dit: Advancing video generation with pseudo 4d gaussian fields through efficient dense 3d point tracking.
\newblock \emph{arXiv preprint arXiv:2501.02690}, 2025.

\bibitem[Blattmann et~al.(2023)Blattmann, Dockhorn, Kulal, Mendelevitch, Kilian, Lorenz, Levi, English, Voleti, Letts, et~al.]{svd}
Andreas Blattmann, Tim Dockhorn, Sumith Kulal, Daniel Mendelevitch, Maciej Kilian, Dominik Lorenz, Yam Levi, Zion English, Vikram Voleti, Adam Letts, et~al.
\newblock Stable video diffusion: Scaling latent video diffusion models to large datasets.
\newblock \emph{arXiv preprint arXiv:2311.15127}, 2023.

\bibitem[Chen et~al.(2023{\natexlab{a}})Chen, Xia, He, Zhang, Cun, Yang, Xing, Liu, Chen, Wang, et~al.]{chen2023videocrafter1}
Haoxin Chen, Menghan Xia, Yingqing He, Yong Zhang, Xiaodong Cun, Shaoshu Yang, Jinbo Xing, Yaofang Liu, Qifeng Chen, Xintao Wang, et~al.
\newblock Videocrafter1: Open diffusion models for high-quality video generation.
\newblock \emph{arXiv preprint arXiv:2310.19512}, 2023{\natexlab{a}}.

\bibitem[Chen et~al.(2023{\natexlab{b}})Chen, Yu, Ge, Yao, Xie, Wu, Wang, Kwok, Luo, Lu, et~al.]{chen2023pixart}
Junsong Chen, Jincheng Yu, Chongjian Ge, Lewei Yao, Enze Xie, Yue Wu, Zhongdao Wang, James Kwok, Ping Luo, Huchuan Lu, et~al.
\newblock Pixart-alpha: Fast training of diffusion transformer for photorealistic text-to-image synthesis.
\newblock \emph{arXiv preprint arXiv:2310.00426}, 2023{\natexlab{b}}.

\bibitem[Chen et~al.(2023{\natexlab{c}})Chen, Ji, Wu, Wu, Xie, Li, Xia, Xiao, and Lin]{chen2023control}
Weifeng Chen, Yatai Ji, Jie Wu, Hefeng Wu, Pan Xie, Jiashi Li, Xin Xia, Xuefeng Xiao, and Liang Lin.
\newblock Control-a-video: Controllable text-to-video generation with diffusion models.
\newblock \emph{arXiv preprint arXiv:2305.13840}, 2023{\natexlab{c}}.

\bibitem[Chen et~al.(2025)Chen, Men, Yao, Cui, and Bo]{chen2025perception}
Yingjie Chen, Yifang Men, Yuan Yao, Miaomiao Cui, and Liefeng Bo.
\newblock Perception-as-control: Fine-grained controllable image animation with 3d-aware motion representation.
\newblock \emph{arXiv preprint arXiv:2501.05020}, 2025.

\bibitem[Dehghani et~al.(2024)Dehghani, Mustafa, Djolonga, Heek, Minderer, Caron, Steiner, Puigcerver, Geirhos, Alabdulmohsin, et~al.]{dehghani2024patch}
Mostafa Dehghani, Basil Mustafa, Josip Djolonga, Jonathan Heek, Matthias Minderer, Mathilde Caron, Andreas Steiner, Joan Puigcerver, Robert Geirhos, Ibrahim~M Alabdulmohsin, et~al.
\newblock Patch n’pack: Navit, a vision transformer for any aspect ratio and resolution.
\newblock \emph{Advances in Neural Information Processing Systems}, 36, 2024.

\bibitem[Esser et~al.(2024)Esser, Kulal, Blattmann, Entezari, M{\"u}ller, Saini, Levi, Lorenz, Sauer, Boesel, et~al.]{esser2024scaling}
Patrick Esser, Sumith Kulal, Andreas Blattmann, Rahim Entezari, Jonas M{\"u}ller, Harry Saini, Yam Levi, Dominik Lorenz, Axel Sauer, Frederic Boesel, et~al.
\newblock Scaling rectified flow transformers for high-resolution image synthesis.
\newblock In \emph{Forty-first International Conference on Machine Learning}, 2024.

\bibitem[Fu et~al.(2024)Fu, Liu, Wang, Peng, Xia, Shi, Yuan, Wan, Zhang, and Lin]{fu20243dtrajmaster}
Xiao Fu, Xian Liu, Xintao Wang, Sida Peng, Menghan Xia, Xiaoyu Shi, Ziyang Yuan, Pengfei Wan, Di Zhang, and Dahua Lin.
\newblock 3dtrajmaster: Mastering 3d trajectory for multi-entity motion in video generation.
\newblock \emph{arXiv preprint arXiv:2412.07759}, 2024.

\bibitem[Geng et~al.(2024)Geng, Herrmann, Hur, Cole, Zhang, Pfaff, Lopez-Guevara, Doersch, Aytar, Rubinstein, et~al.]{geng2024motion}
Daniel Geng, Charles Herrmann, Junhwa Hur, Forrester Cole, Serena Zhang, Tobias Pfaff, Tatiana Lopez-Guevara, Carl Doersch, Yusuf Aytar, Michael Rubinstein, et~al.
\newblock Motion prompting: Controlling video generation with motion trajectories.
\newblock \emph{arXiv preprint arXiv:2412.02700}, 2024.

\bibitem[Gu et~al.(2025)Gu, Yan, Lu, Li, Dou, Si, Dong, Liu, Lin, Liu, et~al.]{shader}
Zekai Gu, Rui Yan, Jiahao Lu, Peng Li, Zhiyang Dou, Chenyang Si, Zhen Dong, Qifeng Liu, Cheng Lin, Ziwei Liu, et~al.
\newblock Diffusion as shader: 3d-aware video diffusion for versatile video generation control.
\newblock \emph{arXiv preprint arXiv:2501.03847}, 2025.

\bibitem[Guo et~al.(2023{\natexlab{a}})Guo, Yang, Rao, Agrawala, Lin, and Dai]{guo2023sparsectrl}
Yuwei Guo, Ceyuan Yang, Anyi Rao, Maneesh Agrawala, Dahua Lin, and Bo Dai.
\newblock Sparsectrl: Adding sparse controls to text-to-video diffusion models, 2023{\natexlab{a}}.

\bibitem[Guo et~al.(2023{\natexlab{b}})Guo, Yang, Rao, Liang, Wang, Qiao, Agrawala, Lin, and Dai]{animatediff}
Yuwei Guo, Ceyuan Yang, Anyi Rao, Zhengyang Liang, Yaohui Wang, Yu Qiao, Maneesh Agrawala, Dahua Lin, and Bo Dai.
\newblock Animatediff: Animate your personalized text-to-image diffusion models without specific tuning.
\newblock \emph{arXiv preprint arXiv:2307.04725}, 2023{\natexlab{b}}.

\bibitem[He et~al.(2024)He, Xu, Guo, Wetzstein, Dai, Li, and Yang]{cameractrl}
Hao He, Yinghao Xu, Yuwei Guo, Gordon Wetzstein, Bo Dai, Hongsheng Li, and Ceyuan Yang.
\newblock Cameractrl: Enabling camera control for text-to-video generation, 2024.

\bibitem[Ho and Salimans(2022)]{ho2022classifier}
Jonathan Ho and Tim Salimans.
\newblock Classifier-free diffusion guidance.
\newblock \emph{arXiv preprint arXiv:2207.12598}, 2022.

\bibitem[Ho et~al.(2020)Ho, Jain, and Abbeel]{ddpm}
Jonathan Ho, Ajay Jain, and Pieter Abbeel.
\newblock Denoising diffusion probabilistic models.
\newblock \emph{Advances in neural information processing systems}, 33:\penalty0 6840--6851, 2020.

\bibitem[Hu(2024)]{hu2024animate}
Li Hu.
\newblock Animate anyone: Consistent and controllable image-to-video synthesis for character animation.
\newblock In \emph{Proceedings of the IEEE/CVF Conference on Computer Vision and Pattern Recognition}, pages 8153--8163, 2024.

\bibitem[Kingma(2013)]{kingma2013auto}
Diederik~P Kingma.
\newblock Auto-encoding variational bayes.
\newblock \emph{arXiv preprint arXiv:1312.6114}, 2013.

\bibitem[Kingma(2014)]{kingma2014adam}
Diederik~P Kingma.
\newblock Adam: A method for stochastic optimization.
\newblock \emph{arXiv preprint arXiv:1412.6980}, 2014.

\bibitem[Koroglu et~al.(2024)Koroglu, Caselles-Dupr{\'e}, Sanmiguel, and Cord]{koroglu2024onlyflow}
Mathis Koroglu, Hugo Caselles-Dupr{\'e}, Guillaume~Jeanneret Sanmiguel, and Matthieu Cord.
\newblock Onlyflow: Optical flow based motion conditioning for video diffusion models.
\newblock \emph{arXiv preprint arXiv:2411.10501}, 2024.

\bibitem[Lin et~al.(2014)Lin, Maire, Belongie, Hays, Perona, Ramanan, Doll{\'a}r, and Zitnick]{lin2014microsoft}
Tsung-Yi Lin, Michael Maire, Serge Belongie, James Hays, Pietro Perona, Deva Ramanan, Piotr Doll{\'a}r, and C~Lawrence Zitnick.
\newblock Microsoft coco: Common objects in context.
\newblock In \emph{Computer Vision--ECCV 2014: 13th European Conference, Zurich, Switzerland, September 6-12, 2014, Proceedings, Part V 13}, pages 740--755. Springer, 2014.

\bibitem[Lipman et~al.(2022)Lipman, Chen, Ben-Hamu, Nickel, and Le]{lipman2022flow}
Yaron Lipman, Ricky~TQ Chen, Heli Ben-Hamu, Maximilian Nickel, and Matt Le.
\newblock Flow matching for generative modeling.
\newblock \emph{arXiv preprint arXiv:2210.02747}, 2022.

\bibitem[Liu et~al.(2023)Liu, Zeng, Ren, Li, Zhang, Yang, Li, Yang, Su, Zhu, et~al.]{groundingdino}
Shilong Liu, Zhaoyang Zeng, Tianhe Ren, Feng Li, Hao Zhang, Jie Yang, Chunyuan Li, Jianwei Yang, Hang Su, Jun Zhu, et~al.
\newblock Grounding dino: Marrying dino with grounded pre-training for open-set object detection.
\newblock \emph{arXiv preprint arXiv:2303.05499}, 2023.

\bibitem[Lucic et~al.(2017)Lucic, Kurach, Michalski, Gelly, and Bousquet]{lucic2017gans}
Mario Lucic, Karol Kurach, Marcin Michalski, Sylvain Gelly, and Olivier Bousquet.
\newblock Are gans created equal? a large-scale study.
\newblock \emph{arXiv preprint arXiv:1711.10337}, 2017.

\bibitem[Mou et~al.(2024)Mou, Wang, Xie, Wu, Zhang, Qi, and Shan]{mou2024t2i}
Chong Mou, Xintao Wang, Liangbin Xie, Yanze Wu, Jian Zhang, Zhongang Qi, and Ying Shan.
\newblock T2i-adapter: Learning adapters to dig out more controllable ability for text-to-image diffusion models.
\newblock In \emph{Proceedings of the AAAI Conference on Artificial Intelligence}, pages 4296--4304, 2024.

\bibitem[Peebles and Xie(2023)]{peebles2023scalable}
William Peebles and Saining Xie.
\newblock Scalable diffusion models with transformers.
\newblock In \emph{Proceedings of the IEEE/CVF International Conference on Computer Vision}, pages 4195--4205, 2023.

\bibitem[Raffel et~al.(2020)Raffel, Shazeer, Roberts, Lee, Narang, Matena, Zhou, Li, and Liu]{raffel2020exploring}
Colin Raffel, Noam Shazeer, Adam Roberts, Katherine Lee, Sharan Narang, Michael Matena, Yanqi Zhou, Wei Li, and Peter~J Liu.
\newblock Exploring the limits of transfer learning with a unified text-to-text transformer.
\newblock \emph{Journal of machine learning research}, 21\penalty0 (140):\penalty0 1--67, 2020.

\bibitem[Ravi et~al.(2024)Ravi, Gabeur, Hu, Hu, Ryali, Ma, Khedr, Rädle, Rolland, Gustafson, Mintun, Pan, Alwala, Carion, Wu, Girshick, Dollár, and Feichtenhofer]{2024sam2segmentimages}
Nikhila Ravi, Valentin Gabeur, Yuan-Ting Hu, Ronghang Hu, Chaitanya Ryali, Tengyu Ma, Haitham Khedr, Roman Rädle, Chloe Rolland, Laura Gustafson, Eric Mintun, Junting Pan, Kalyan~Vasudev Alwala, Nicolas Carion, Chao-Yuan Wu, Ross Girshick, Piotr Dollár, and Christoph Feichtenhofer.
\newblock Sam 2: Segment anything in images and videos, 2024.

\bibitem[Rombach et~al.(2022)Rombach, Blattmann, Lorenz, Esser, and Ommer]{sd}
Robin Rombach, Andreas Blattmann, Dominik Lorenz, Patrick Esser, and Bj{\"o}rn Ommer.
\newblock High-resolution image synthesis with latent diffusion models.
\newblock In \emph{Proceedings of the IEEE/CVF conference on computer vision and pattern recognition}, pages 10684--10695, 2022.

\bibitem[Shao et~al.(2019)Shao, Li, Zhang, Peng, Yu, Zhang, Li, and Sun]{shao2019objects365}
Shuai Shao, Zeming Li, Tianyuan Zhang, Chao Peng, Gang Yu, Xiangyu Zhang, Jing Li, and Jian Sun.
\newblock Objects365: A large-scale, high-quality dataset for object detection.
\newblock In \emph{Proceedings of the IEEE/CVF international conference on computer vision}, pages 8430--8439, 2019.

\bibitem[Shi et~al.(2024)Shi, Huang, Wang, Bian, Li, Zhang, Zhang, Cheung, See, Qin, et~al.]{motioni2v}
Xiaoyu Shi, Zhaoyang Huang, Fu-Yun Wang, Weikang Bian, Dasong Li, Yi Zhang, Manyuan Zhang, Ka~Chun Cheung, Simon See, Hongwei Qin, et~al.
\newblock Motion-i2v: Consistent and controllable image-to-video generation with explicit motion modeling.
\newblock In \emph{ACM SIGGRAPH 2024 Conference Papers}, pages 1--11, 2024.

\bibitem[Shuai et~al.(2025)Shuai, Ding, Qin, Luo, Ma, and Tao]{shuai2025free}
Xincheng Shuai, Henghui Ding, Zhenyuan Qin, Hao Luo, Xingjun Ma, and Dacheng Tao.
\newblock Free-form motion control: A synthetic video generation dataset with controllable camera and object motions.
\newblock \emph{arXiv preprint arXiv:2501.01425}, 2025.

\bibitem[Song et~al.(2020{\natexlab{a}})Song, Meng, and Ermon]{ddim}
Jiaming Song, Chenlin Meng, and Stefano Ermon.
\newblock Denoising diffusion implicit models.
\newblock \emph{arXiv preprint arXiv:2010.02502}, 2020{\natexlab{a}}.

\bibitem[Song et~al.(2020{\natexlab{b}})Song, Meng, and Ermon]{song2020denoising}
Jiaming Song, Chenlin Meng, and Stefano Ermon.
\newblock Denoising diffusion implicit models.
\newblock \emph{arXiv preprint arXiv:2010.02502}, 2020{\natexlab{b}}.

\bibitem[Unterthiner et~al.(2018)Unterthiner, Van~Steenkiste, Kurach, Marinier, Michalski, and Gelly]{unterthiner2018towards}
Thomas Unterthiner, Sjoerd Van~Steenkiste, Karol Kurach, Raphael Marinier, Marcin Michalski, and Sylvain Gelly.
\newblock Towards accurate generative models of video: A new metric \& challenges.
\newblock \emph{arXiv preprint arXiv:1812.01717}, 2018.

\bibitem[Wang et~al.(2024{\natexlab{a}})Wang, Gu, Hu, Zhao, Guo, Han, Xu, and Liang]{wang2024easycontrol}
Cong Wang, Jiaxi Gu, Panwen Hu, Haoyu Zhao, Yuanfan Guo, Jianhua Han, Hang Xu, and Xiaodan Liang.
\newblock Easycontrol: Transfer controlnet to video diffusion for controllable generation and interpolation.
\newblock \emph{arXiv preprint arXiv:2408.13005}, 2024{\natexlab{a}}.

\bibitem[Wang et~al.(2024{\natexlab{b}})Wang, Zhang, Zou, Zeng, Wei, Yuan, and Li]{boximator}
Jiawei Wang, Yuchen Zhang, Jiaxin Zou, Yan Zeng, Guoqiang Wei, Liping Yuan, and Hang Li.
\newblock Boximator: Generating rich and controllable motions for video synthesis, 2024{\natexlab{b}}.

\bibitem[Wang et~al.(2023)Wang, Yuan, Zhang, Chen, Wang, Zhang, Shen, Zhao, and Zhou]{VideoComposer}
Xiang Wang, Hangjie Yuan, Shiwei Zhang, Dayou Chen, Jiuniu Wang, Yingya Zhang, Yujun Shen, Deli Zhao, and Jingren Zhou.
\newblock {VideoComposer}: {Compositional} {Video} {Synthesis} with {Motion} {Controllability}, 2023.
\newblock arXiv:2306.02018 [cs].

\bibitem[Wang et~al.(2024{\natexlab{c}})Wang, Yuan, Wang, Chen, Xia, Luo, and Shan]{MotionCtrl}
Zhouxia Wang, Ziyang Yuan, Xintao Wang, Tianshui Chen, Menghan Xia, Ping Luo, and Ying Shan.
\newblock {MotionCtrl}: {A} {Unified} and {Flexible} {Motion} {Controller} for {Video} {Generation}, 2024{\natexlab{c}}.
\newblock arXiv:2312.03641 [cs].

\bibitem[Xiao et~al.(2024)Xiao, Wang, Zhang, Xue, Peng, Shen, and Zhou]{SpatialTracker}
Yuxi Xiao, Qianqian Wang, Shangzhan Zhang, Nan Xue, Sida Peng, Yujun Shen, and Xiaowei Zhou.
\newblock Spatialtracker: Tracking any 2d pixels in 3d space.
\newblock In \emph{Proceedings of the IEEE/CVF Conference on Computer Vision and Pattern Recognition (CVPR)}, 2024.

\bibitem[Xing et~al.(2025)Xing, Mai, Ham, Huang, Mahapatra, Fu, Wong, and Liu]{motioncanvas}
Jinbo Xing, Long Mai, Cusuh Ham, Jiahui Huang, Aniruddha Mahapatra, Chi-Wing Fu, Tien-Tsin Wong, and Feng Liu.
\newblock Motioncanvas: Cinematic shot design with controllable image-to-video generation, 2025.

\bibitem[Yang et~al.(2024{\natexlab{a}})Yang, Yang, Hui, Zheng, Yu, Zhou, Li, Li, Liu, Huang, Dong, Wei, Lin, Tang, Wang, Yang, Tu, Zhang, Ma, Xu, Zhou, Bai, He, Lin, Dang, Lu, Chen, Yang, Li, Xue, Ni, Zhang, Wang, Peng, Men, Gao, Lin, Wang, Bai, Tan, Zhu, Li, Liu, Ge, Deng, Zhou, Ren, Zhang, Wei, Ren, Fan, Yao, Zhang, Wan, Chu, Liu, Cui, Zhang, and Fan]{qwen2}
An Yang, Baosong Yang, Binyuan Hui, Bo Zheng, Bowen Yu, Chang Zhou, Chengpeng Li, Chengyuan Li, Dayiheng Liu, Fei Huang, Guanting Dong, Haoran Wei, Huan Lin, Jialong Tang, Jialin Wang, Jian Yang, Jianhong Tu, Jianwei Zhang, Jianxin Ma, Jin Xu, Jingren Zhou, Jinze Bai, Jinzheng He, Junyang Lin, Kai Dang, Keming Lu, Keqin Chen, Kexin Yang, Mei Li, Mingfeng Xue, Na Ni, Pei Zhang, Peng Wang, Ru Peng, Rui Men, Ruize Gao, Runji Lin, Shijie Wang, Shuai Bai, Sinan Tan, Tianhang Zhu, Tianhao Li, Tianyu Liu, Wenbin Ge, Xiaodong Deng, Xiaohuan Zhou, Xingzhang Ren, Xinyu Zhang, Xipin Wei, Xuancheng Ren, Yang Fan, Yang Yao, Yichang Zhang, Yu Wan, Yunfei Chu, Yuqiong Liu, Zeyu Cui, Zhenru Zhang, and Zhihao Fan.
\newblock Qwen2 technical report.
\newblock \emph{arXiv preprint arXiv:2407.10671}, 2024{\natexlab{a}}.

\bibitem[Yang et~al.(2024{\natexlab{b}})Yang, Kang, Huang, Zhao, Xu, Feng, and Zhao]{depth_anything_v2}
Lihe Yang, Bingyi Kang, Zilong Huang, Zhen Zhao, Xiaogang Xu, Jiashi Feng, and Hengshuang Zhao.
\newblock Depth anything v2.
\newblock \emph{arXiv:2406.09414}, 2024{\natexlab{b}}.

\bibitem[Yang et~al.(2024{\natexlab{c}})Yang, Hou, Huang, Ma, Wan, Zhang, Chen, and Liao]{yang2024direct}
Shiyuan Yang, Liang Hou, Haibin Huang, Chongyang Ma, Pengfei Wan, Di Zhang, Xiaodong Chen, and Jing Liao.
\newblock Direct-a-video: Customized video generation with user-directed camera movement and object motion.
\newblock In \emph{ACM SIGGRAPH 2024 Conference Papers}, pages 1--12, 2024{\natexlab{c}}.

\bibitem[Yin et~al.(2023)Yin, Wu, Liang, Shi, Li, Ming, and Duan]{dragnuwa}
Shengming Yin, Chenfei Wu, Jian Liang, Jie Shi, Houqiang Li, Gong Ming, and Nan Duan.
\newblock Dragnuwa: Fine-grained control in video generation by integrating text, image, and trajectory, 2023.

\bibitem[Zhang et~al.(2024{\natexlab{a}})Zhang, Herrmann, Hur, Jampani, Darrell, Cole, Sun, and Yang]{zhang2024monst3r}
Junyi Zhang, Charles Herrmann, Junhwa Hur, Varun Jampani, Trevor Darrell, Forrester Cole, Deqing Sun, and Ming-Hsuan Yang.
\newblock Monst3r: A simple approach for estimating geometry in the presence of motion.
\newblock \emph{arXiv preprint arxiv:2410.03825}, 2024{\natexlab{a}}.

\bibitem[Zhang et~al.(2023)Zhang, Rao, and Agrawala]{controlnet}
Lvmin Zhang, Anyi Rao, and Maneesh Agrawala.
\newblock Adding conditional control to text-to-image diffusion models, 2023.

\bibitem[Zhang et~al.(2024{\natexlab{b}})Zhang, Liao, Li, Qin, and Wang]{Tora2024}
Zhenghao Zhang, Junchao Liao, Menghao Li, Long Qin, and Weizhi Wang.
\newblock Tora: {Trajectory}-oriented {Diffusion} {Transformer} for {Video} {Generation}, 2024{\natexlab{b}}.
\newblock arXiv:2407.21705 [cs].

\bibitem[Zhou et~al.(2018)Zhou, Tucker, Flynn, Fyffe, and Snavely]{realestate10k}
Tinghui Zhou, Richard Tucker, John Flynn, Graham Fyffe, and Noah Snavely.
\newblock Stereo magnification: Learning view synthesis using multiplane images, 2018.

\end{thebibliography}

\end{document}